\documentclass[letterpaper, 10 pt, conference]{ieeeconf}

\IEEEoverridecommandlockouts
\usepackage[T1]{fontenc}          
\usepackage{graphicx}
\usepackage{amsmath,amssymb}
\usepackage{booktabs}             
\usepackage{multirow}
\usepackage{xcolor}
\usepackage{capt-of}

\newcommand{\method}{STVLA}                  

\newcommand{\base}{$\pi_{0.5}$}              
\newcommand{\geo}{VGGT-$\Omega$}                
\newcommand{\compressor}{scene-token compressor}

\newcommand{\vtok}{\mathbf{V}}               
\newcommand{\stok}{\mathbf{S}}               
\newcommand{\mtok}{\mathbf{M}}               
\newcommand{\hist}{\mathcal{H}}              

\usepackage[hidelinks]{hyperref}
\renewcommand{\footnoterule}{\kern -3pt\hrule width 0.25\columnwidth\kern 2.6pt}

\newif\ifappendix \appendixfalse
\ifdefined\WITHAPPENDIX \appendixtrue \fi

\title{\LARGE \bf
STVLA: Spatial-aware and Temporal-memory for  Vision-Language-Action Models
}
\author{\authorblockN{\bfseries Zhuo Han, Yeqiang Wang, Xinran Wang, Hongwei Han, Yi Xiong, Zunian Wan,\\
 Chongqing Huang\authorrefmark{3}, Jianbo Mao, Huan Peng, Yongcheng Liu, Wei Shan, Maoqing Yao\authorrefmark{2}}\\[-0.8ex]
\authorblockA{AgiBot}\\[-0.8ex]
\thanks{\authorrefmark{2} Corresponding author: Maoqing Yao.}%
\thanks{\authorrefmark{3} Project lead: Chongqing Huang.}}

\begin{document}
\IEEEaftertitletext{%
  \vspace{-1\baselineskip}
  \begin{minipage}{\textwidth}
    \centering
    \includegraphics[width=\textwidth]{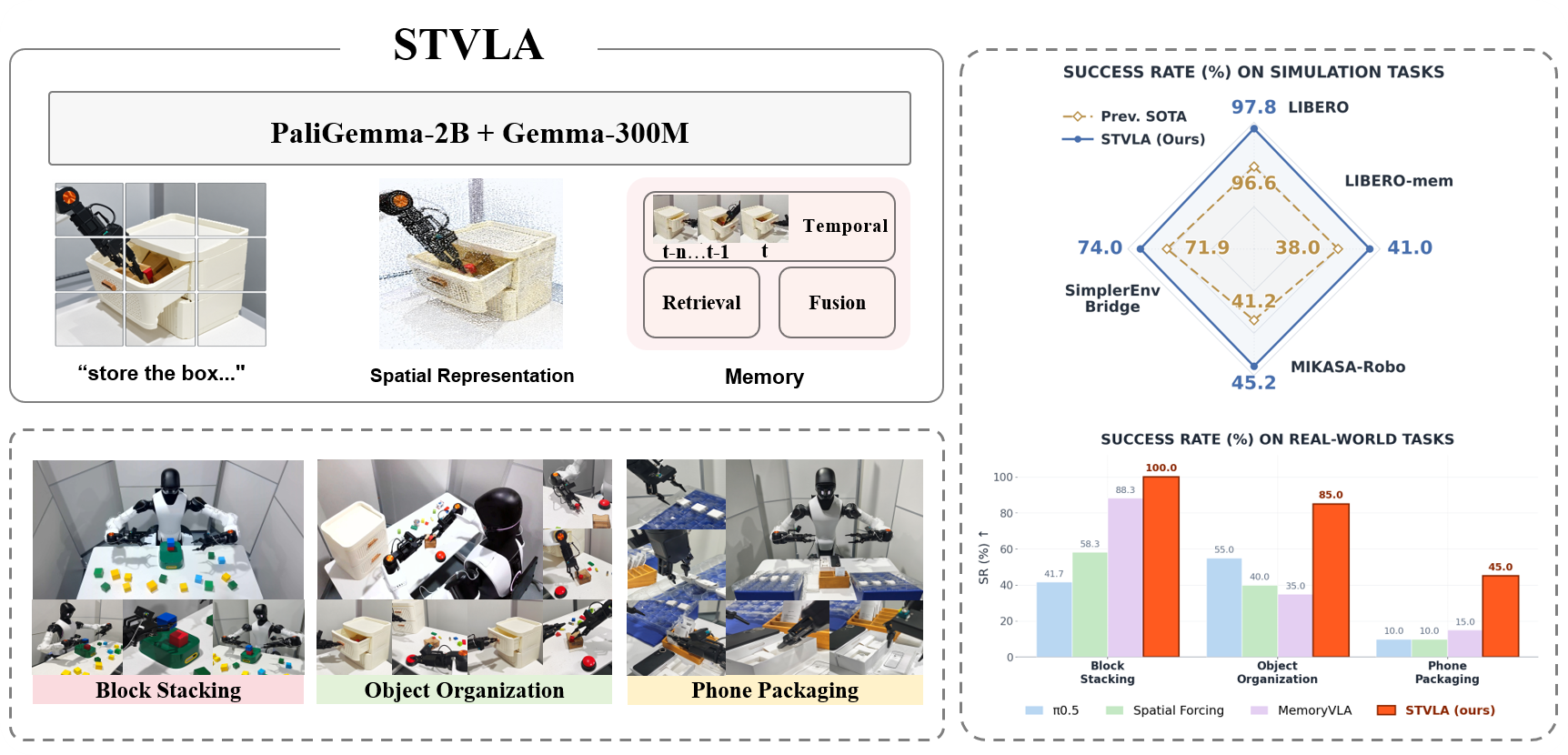}
    \captionof{figure}{\textbf{We present \method{}, a spatial-aware and temporal-memory vision-language-action model for robotic manipulation.}
    \method{} combines compact spatial representations distilled from a frozen visual geometry model with a bounded visual--spatial memory, enabling the policy to retrieve and fuse task-relevant historical evidence during action generation. Across real-world block stacking, object organization, and phone packaging, the tasks collectively stress geometry-sensitive manipulation and progress tracking through multi-step execution. Across four simulated benchmarks and three real-world manipulation tasks, \method{} consistently outperforms strong VLA, memory-augmented, and spatially aware baselines.}
    \label{fig:teaser}
  \end{minipage}
}
\maketitle
v\thispagestyle{empty}
\pagestyle{empty}



\begin{abstract}
Robotic manipulation often requires both spatial scene understanding and task history. However, many vision-language-action (VLA) policies rely primarily on current observations, without explicitly integrating spatial structure and task history. Appearance-based memories can be sensitive to occlusion and viewpoint changes, while spatial representations of the current scene alone do not preserve how the scene has changed during task execution. We introduce STVLA, a framework that incorporates features from a pretrained visual geometry foundation model into temporal memory alongside visual representations. This allows the policy to recall past scene structure as context for current action decisions. 
To select informative history, STVLA uses a mathematically grounded scoring rule that balances relevance to the current observation with complementary evidence from memory.
Across evaluations on LIBERO, LIBERO-Mem, MIKASA-Robo, and SimplerEnv Bridge, STVLA achieves the highest average success rate among the compared methods on each benchmark. On real-world block stacking, object organization, and phone packaging, it improves success over $\pi_{0.5}$ by $58.33$, $30.00$, and $35.00$ percentage points, respectively. These results support integrating scene geometry into temporal memory to improve manipulation requiring both spatial precision and task history.
Project website: \href{https://stvla.github.io/}{https://stvla.github.io/}.
\end{abstract}

\section{INTRODUCTION}
\label{sec}

Vision-language-action (VLA) policies have achieved strong performance in robotic manipulation \cite{openvla,pi0,pi05}, yet many condition each action primarily on the current observation, leaving both task history and spatial scene structure only implicitly represented. Everyday manipulation tasks are often non-Markovian from the policy's perspective: visually similar observations can correspond to different task states, causing a current-frame policy to repeat completed steps or stall. When the action further depends on the geometric structure of the scene --- the height of a surface, whether a location is still free --- 2D features extracted from a single frame and a single view do not express that structure. In everyday manipulation, the two demands often arise together.

Memory-augmented policies retain information from previous observations to support decisions over longer horizons \cite{hiveformer,vima,eventvla,kemo,memoryvla}, while spatially aware policies enrich visual representations with depth, 3D structure, or features learned by geometry models \cite{threedvla,spatialvla,spatialforcing}. These capabilities, however, are commonly introduced as separate components. A memory built on appearance features inherits the weaknesses of appearance itself: once the gripper
occludes the workspace or the viewpoint changes, the current frame no longer resembles the stored frame that shares its physical configuration, and retrieval fails in exactly the situations where memory should matter most. A spatial representation computed on the current observation alone has the mirror-image problem: it describes what the scene is, but not which parts of it the policy has already changed. This motivates a simple principle: the spatial representation of the scene should be part of what the policy remembers.

We introduce \method{}, a VLA architecture that integrates spatial
representations directly into bounded temporal memory(Fig.~\ref{fig:teaser}). A frozen pretrained visual geometry model \cite{wang2026vggtomega} extracts scene geometry features from a multi-frame window of RGB observations, and a trainable compressor maps its intermediate features into compact spatial tokens. These spatial tokens are paired with the backbone's visual tokens and stored in a single bounded memory bank. At each step, the current visual and spatial representations jointly retrieve relevant information from the stored history through cross-attention. Retrieved spatial information conditions the action expert through a dedicated prefix segment. This design allows spatial information and temporal context to interact during memory retrieval while preserving the backbone's pretrained image-language representation.

Our contributions are summarized as follows:
\begin{itemize}
\setlength{\itemsep}{1pt}
\item We introduce a memory paradigm for VLA policies that incorporates
features from a pretrained visual geometry foundation model into temporal
memory. This makes past scene structure available for action generation,
enabling the policy to use both appearance and geometry as historical context.

\item We develop a joint visual-spatial retrieval mechanism for conditioning
action generation on useful history. A relevance--non-redundancy score, computed from
current--memory feature similarity and normalized feature difference, serves as an entry-level
attention prior without introducing an auxiliary learned scorer.
\item We evaluate \method{} on four simulation benchmarks and three real-world
tasks. \method{} improves average success over our
$\pi_{0.5}$ baseline on each simulation benchmark and improves success by
$30.0$--$58.3$ percentage points on real-world tasks that jointly stress
geometry-sensitive manipulation and progress tracking.
\end{itemize}

\section{RELATED WORK}
\label{sec:related}

\subsection{Vision-Language-Action Policies}

Recent VLA models extend pretrained vision-language models (VLMs) to robot action generation. OpenVLA~\cite{openvla} discretize robot actions and predict them autoregressively as tokens. In contrast, $\pi_0$ and $\pi_{0.5}$~\cite{pi0,pi05} condition dedicated continuous-action modules on VLM representations and generate action chunks through flow matching.

\subsection{Memory for Robot Policies}

Memory-augmented robot policies retain past information through different history-selection and memory-management mechanisms. Early dense-history methods append encoded observations and actions to the Transformer context: Hiveformer~\cite{hiveformer} retains the full multi-view history, while VIMA~\cite{vima} models interleaved observation--action tokens; their sequence length and attention cost grow with retained history. Sparse keyframe-memory methods reduce the retained context: EventVLA~\cite{eventvla} predicts future keyframe probabilities from VLA hidden states, whereas KEMO~\cite{kemo} identifies candidate events from kinematic cues and filters them by visual change using a frozen encoder. Rather than selecting only keyframes, MemoryVLA~\cite{memoryvla} maintains a bounded Perceptual--Cognitive Memory Bank, retrieves task-relevant entries, and, when full, merges the most similar temporally adjacent pair in each perceptual and cognitive stream.

\subsection{Spatial Reasoning for Manipulation}

Recent works introduce spatial information into robot policies through different representations and architectures. 3D-VLA~\cite{threedvla} links 3D perception, reasoning, and action through a 3D generative world model. SpatialVLA~\cite{spatialvla} injects Ego3D position encodings into VLA observations, while DepthVLA~\cite{depthvla} incorporates a pretrained depth transformer alongside the VLM and action expert. Spatial Forcing~\cite{spatialforcing} aligns intermediate VLA features with representations from a pretrained 3D foundation model without requiring explicit 3D inputs at inference. These methods strengthen spatial or spatiotemporal reasoning; \method{} differs by storing paired appearance and explicit 3D scene tokens as retrievable per-episode memory and exposing the recalled geometry directly to the VLA action pathway.

\section{METHOD}
\label{sec:method}

\begin{figure*}[t]
  \vspace*{4pt}
  \centering
  \includegraphics[width=\textwidth]{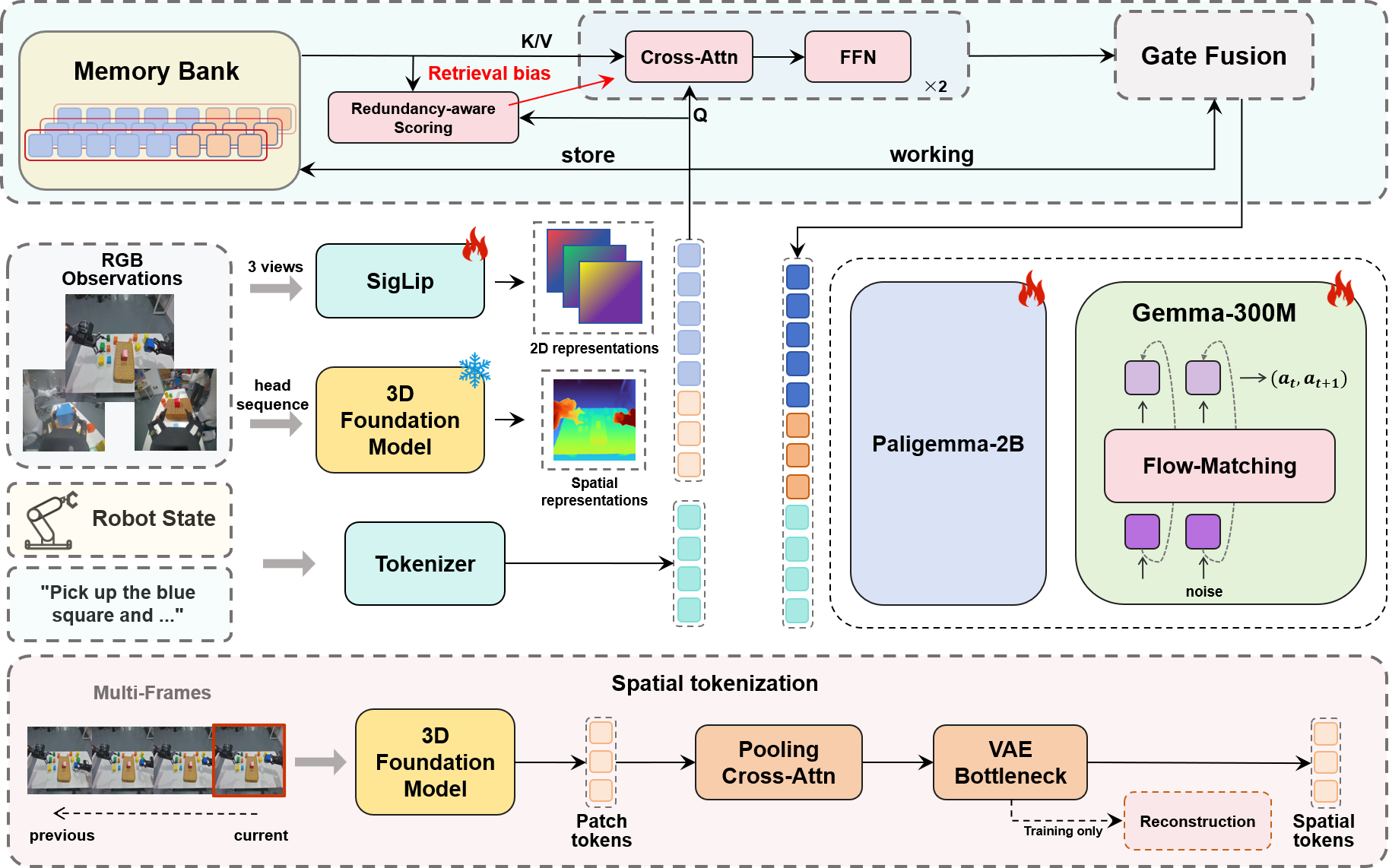}
  \caption{\textbf{Architecture of \method{}.}
    Current visual and spatial tokens query a bounded memory bank through score-biased cross-attention (top). Retrieved features are gated into the current representations to condition action generation (middle). The spatial tokenization pathway (bottom) compresses current-frame features from frozen \geo{} through pooling cross-attention and a variational bottleneck, with an auxiliary reconstruction branch used only during training. Flame and snowflake icons denote trainable and frozen modules, respectively.}
  \label{fig:arch}
  \vspace{-10pt}
\end{figure*}

\subsection{Architecture overview}
\label{sec:method:overview}

As illustrated in Fig.~\ref{fig:arch}, \method{} augments the PaliGemma-based vision--language model~\cite{paligemma} and flow-matching action expert of \base{}~\cite{pi05} with spatial tokenization and bounded visual--spatial memory. Current observations are encoded into visual tokens by SigLIP~\cite{siglip} and compact spatial tokens through a frozen geometry model and a trainable compressor. These tokens jointly retrieve historical visual--spatial features through score-biased cross-attention. Gated fusion combines the retrieved features with the current representations, which then condition action generation through separate policy-prefix segments alongside the task instruction and robot state.

\subsection{Spatial tokenization}
\label{sec:method:spatial}
We leverage frozen \geo{}~\cite{wang2026vggtomega} to support spatial scene understanding in the policy, complementing the backbone's 2D visual representations. Given a window of RGB observations, the model uses preceding frames as context to extract spatial features of the current scene. Only the features corresponding to the current frame are passed to the compressor.

To incorporate these features within the policy-prefix token budget, a trainable compressor uses learned-query cross-attention to aggregate patch features into compact latent representations. A variational bottleneck regularizes these representations before projection to the language-model width, producing $N_s$ spatial tokens $\stok_t$. The compressor is jointly trained with the policy under the objectives defined in Sec.~\ref{sec:method:loss}.

\begin{figure}[t]
  \centering
  \includegraphics[width=0.6\columnwidth]{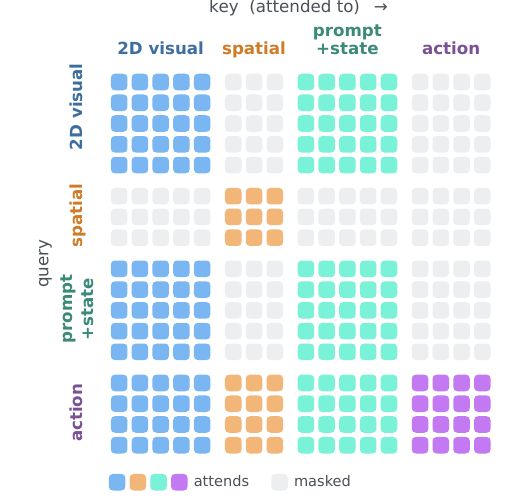}
  \caption{\textbf{Block-causal attention mask over four segments.}
  Rows denote queries and columns denote keys; coloured cells indicate permitted attention, while grey cells are masked. Block widths are schematic.}
  \label{fig:mask}
  \vspace{-12pt}
\end{figure}

Spatial tokens occupy a dedicated prefix segment under the block-causal attention mask in Fig.~\ref{fig:mask}. Visual and instruction--state tokens attend bidirectionally within a shared block, while spatial tokens attend only within their own segment and are masked from the original prefix tokens. Action tokens attend to all prefix tokens and all tokens within the action chunk. This preserves the original prefix attention pattern while making spatial information directly accessible to action generation.

\subsection{Bounded temporal memory}
\label{sec:method:stm}

We maintain a bounded memory of paired visual and spatial representations, allowing the policy to retrieve appearance and geometry cues from past observations jointly. At step $t$, let $\mathbf X_t=[\vtok_t;\stok_t]$ concatenate the current visual tokens $\vtok_t$ and spatial tokens $\stok_t$ before memory fusion. The bank contains entries from earlier steps in the same episode:
\begin{equation}
  \label{eq:bank}
  \hist_t
  = (\mathbf X_k)_{k\in\mathcal I_t},
  \qquad |\mathcal I_t|\leq L_m,
\end{equation}
where $\mathcal I_t$ indexes retained steps with $k<t$, and $L_m$ is the memory capacity. The current tokens query the bank through $L$ stacked cross-attention layers:
\begin{equation}
  \label{eq:retrieval}
  \mtok_t =\operatorname{CrossAttn}^{(L)}
\left(
\mathbf X_t,
\mathcal H_t
\right).
\end{equation}
Each query can attend to both visual and spatial tokens across historical entries, enabling cross-modal retrieval rather than restricting each modality to its own history. A learned element-wise gate~\cite{memoryvla} combines the retrieved features with the current tokens. The fused outputs retain separate visual and spatial token groups for policy conditioning. After retrieval, $\mathbf X_t$ is appended to the bank, and the oldest entry is discarded when capacity is exceeded. With no history available, the current tokens pass through unchanged.

\subsection{Relevance and non-redundancy scoring for temporal memory retrieval}
\label{sec:method:retrieval}

To favor history that is relevant to the current observation yet provides complementary information, we introduce a mathematically grounded relevance--non-redundancy score. Inspired by relevance--redundancy criteria in feature selection~\cite{peng2005mrmr}, this non-learned criterion combines similarity and normalized feature difference relative to the current representation without introducing additional learned parameters.

Let $\bar{\mathbf x}_t$ and $\mathbf h_j$ denote the token-wise means of the current tokens and the $j$-th memory entry, respectively.
For nonzero summaries, define $s_j=r_j u_j\in[0,1]$ using
\begin{equation}
  \label{eq:retrieval-score}
  \begin{aligned}
    r_j&=\tfrac12\bigl(1+\cos(\mathbf h_j,\bar{\mathbf x}_t)\bigr),\\
    u_j&=\frac{\lVert\mathbf h_j-\bar{\mathbf x}_t\rVert_2}
      {\lVert\mathbf h_j\rVert_2+\lVert\bar{\mathbf x}_t\rVert_2}.
  \end{aligned}
\end{equation}
Here, $r_j$ measures feature similarity, while $u_j$ measures difference from the current representation. Their product serves as a feature-space proxy for relevance and non-redundancy, rather than a direct estimate of task-specific information gain.

We standardize the scores across retained entries and use them to bias token-level attention:
\begin{equation}
  \label{eq:retrieval-bias}
  \begin{aligned}
    \widetilde{s}_j&=\frac{s_j-\mu_s}
      {\tau_{\mathrm r}\max(\sigma_s,\epsilon)},\\
    \alpha_{j,n}&=\operatorname{softmax}_{j,n}\!\left(
      \frac{\mathbf q^\top\mathbf k_{j,n}}{\sqrt d}
      +\beta\widetilde{s}_j\right).
  \end{aligned}
\end{equation}
Here, $\mu_s$ and $\sigma_s$ are the mean and standard deviation of the entry scores, and $\epsilon>0$ ensures numerical stability. The temperature $\tau_{\mathrm r}>0$ and scale $\beta\geq0$ control the bias strength; $\mathbf q,\mathbf k_{j,n}\in\mathbb R^d$ are the per-head query and key for token $n$ in entry $j$. The softmax covers all tokens in the retained bank. Each entry receives a shared bias, while token-level matching remains query-dependent, guiding retrieval without excluding historical entries.

\subsection{Training objective}
\label{sec:method:loss}
We jointly train the policy and spatial compressor with a flow-matching action objective and auxiliary reconstruction and KL losses:
\begin{equation}
  \label{eq:loss}
  \mathcal{L} \;=\; \mathcal{L}_{\text{act}}
    \;+\; \lambda_{r}\,\mathcal{L}_{\text{recon}}
    \;+\; \lambda_{kl}\,\mathcal{L}_{\text{KL}} .
\end{equation}
For a target action chunk $\mathbf a_t^\star$ of horizon $H$, the action objective follows the backbone's flow-matching formulation~\cite{pi05,flowmatching}. Given the interpolated action $\mathbf x_t^\tau=\tau\boldsymbol\epsilon+(1-\tau)\mathbf a_t^\star$, with $\boldsymbol\epsilon\sim\mathcal N(0,I)$ and $\tau$ sampled according to the backbone's training schedule, we minimize
\begin{equation}
\mathcal{L}_{\mathrm{act}}
=
\mathbb{E}_{\mathbf{a}_t^\star,\boldsymbol{\epsilon},\tau}
\left[
\frac{1}{d_a}
\left\|
\mathbf{v}_\theta
\left(
\mathbf{x}_t^\tau,\tau,\mathbf{c}_t
\right)
-
\left(
\boldsymbol{\epsilon}-\mathbf{a}_t^\star
\right)
\right\|_2^2
\right],
\end{equation}
where $\mathbf c_t$ is the memory-augmented policy context and $d_a$ is the action-chunk dimensionality. This objective also supervises the spatial compressor through the policy-conditioning pathway.

The auxiliary objectives encourage the compressed tokens to retain geometric information. For each spatial token $i$, the compressor predicts a Gaussian posterior $q_{t,i}=\mathcal N(\boldsymbol\mu_{t,i},\operatorname{diag}(\exp\boldsymbol\ell_{t,i}))$. A reparameterized sample is decoded into a reconstructed feature $\hat{\boldsymbol\omega}_{t,i}$. Its target $\bar{\boldsymbol\omega}_{t,i}$ is obtained by fixed average pooling of the frozen geometry features along the patch sequence. We optimize
\begin{equation}
\mathcal{L}_{\mathrm{recon}}
=
\frac{1}{N_s d_\omega}
\sum_{i=1}^{N_s}
\left\|
\hat{\boldsymbol\omega}_{t,i}
-
\bar{\boldsymbol\omega}_{t,i}
\right\|_2^2,
\end{equation}
\begin{equation}
\mathcal{L}_{\mathrm{KL}}
=
\frac{1}{N_s d_z}
\sum_{i=1}^{N_s}
D_{\mathrm{KL}}\!\left(q_{t,i}\,\|\,\mathcal N(0,I)\right),
\end{equation}
where $d_\omega$ and $d_z$ denote the reconstruction-feature and latent dimensions. Reconstruction provides geometric supervision in feature space, while KL regularization constrains the latent posterior toward a standard Gaussian. At inference, the posterior mean is projected into spatial tokens in place of sampling.

%
%
\section{EXPERIMENTS}
\label{sec:experiments}

\subsection{Setup}
\label{sec:exp:setup}
We evaluate \method{} on four simulation benchmarks: LIBERO~\cite{libero}, LIBERO-Mem~\cite{liberomem},
MIKASA-Robo~\cite{mikasa}, and SimplerEnv Bridge~\cite{simplerenv,bridgedatav2}, as well as real-world manipulation tasks.
We report task success rates.

Across both simulation and real-world experiments, we train with $8$ NVIDIA H100 GPUs and a batch size of $16$ per GPU. The VLA image inputs are resized to $224\times224$ during training. We use the frozen 1B-parameter \geo{} with $512$\,px head-camera inputs, yielding $1024$ patch features of width $2048$ per frame. The \compressor{} uses one learned-query pooling cross-attention layer to map these features to $N_s=256$ spatial tokens at the language-model width. For scene reconstruction, \geo{} takes $K=4$ frames from the fixed head-camera view by default. \method{} removes the oldest entries when the memory bank exceeds its capacity and resets the bank at every episode boundary. Across all experiments, \method{} uses a memory capacity of $L_m=16$, a retrieval depth of $L=2$, and a sampler group size of $16$. The auxiliary objective uses $\lambda_r=1$ and $\lambda_{kl}=10^{-4}$.

\subsection{Simulation Experiments}
\label{sec:exp:main}
In simulation, the interval between successive sampled frames matches the action-chunk size
used during evaluation, and we train \base{} and \method{} with the same data and number of optimization steps.

\begin{table}[t]
  \centering
  \caption{\textbf{Success rate (\%) on the five LIBERO suites.}
  Every row above the rule is a prior method; \method{} is ours. Best per column in
  \textbf{bold}.}
  \label{tab:main}
  \setlength{\tabcolsep}{2.5pt}
  {\scriptsize
  \begin{tabular}{lcccccc}
    \toprule
    Method & Spat. & Obj. & Goal & Libero-10 & Libero-90 & Avg. \\
    \midrule
    Diffusion Policy~\cite{diffusionpolicy} & 78.3 & 92.5 & 68.3 & 50.5 & --   & 72.4 \\
    Octo~\cite{octo}                        & 78.9 & 85.7 & 84.6 & 51.1 & --   & 75.1 \\
    UniAct~\cite{uniact}                    & 77.0 & 87.0 & 77.0 & 70.0 & 73.0 & 76.8 \\
    SpatialVLA~\cite{spatialvla}            & 88.2 & 89.9 & 78.6 & 55.5 & 46.2 & 71.7 \\
    TraceVLA~\cite{tracevla}                & 84.6 & 85.2 & 75.1 & 54.1 & --   & 74.8 \\
    OpenVLA~\cite{openvla}                  & 84.7 & 88.4 & 79.2 & 53.7 & 73.5 & 75.9 \\
    CoT-VLA~\cite{cotvla}                   & 87.5 & 91.6 & 87.6 & 69.0 & --   & 81.1 \\
    $\pi_0$-FAST$^{*}$~\cite{pi0fast}       & 96.4 & 96.8 & 88.6 & 60.2 & 83.1 & 85.0 \\
    CronusVLA~\cite{cronusvla}              & 90.1 & 94.7 & 91.3 & 68.7 & --   & 86.2 \\
    CogACT~\cite{cogact}                    & 97.2 & 98.0 & 90.2 & 88.8 & 92.1 & 93.2 \\
    $\pi_0^{*}$~\cite{pi0}                  & 96.8 & 98.8 & 95.8 & 85.2 & --   & 94.2 \\
    MemoryVLA~\cite{memoryvla}              & 98.4 & 98.4 & 96.4 & 93.4 & 95.6 & 96.5 \\
    $\pi_{0.5}^{*\dagger}$~\cite{pi05}      & 97.4 & \textbf{99.2} & 94.6 & 95.4 & 96.4 & 96.6 \\
    \midrule
    \textbf{\method{}}$^{*\dagger}$ \textbf{(ours)}
      & \textbf{98.8} & 98.2 & \textbf{97.6} & \textbf{97.6} & \textbf{96.8}
      & \textbf{97.8}\,\textbf{(+1.2)} \\
    \bottomrule
  \end{tabular}}

  \par\vspace{2pt}
  {\scriptsize\raggedright $^{*}$ denotes additional proprioceptive and wrist-camera inputs.
  $^{\dagger}$ denotes results obtained by us; other results are taken from
  MemoryVLA~\cite{memoryvla} under their respective evaluation settings.
  Reported averages cover five suites when LIBERO-90 is available and four otherwise,
  and are not directly comparable across these groups.
  $+1.2$ denotes the gain over our $\pi_{0.5}$ baseline.\par}
  \vspace{-12pt}
\end{table}


\begin{figure*}[t]
  \vspace*{2pt}
  \centering
  \includegraphics[width=\textwidth]{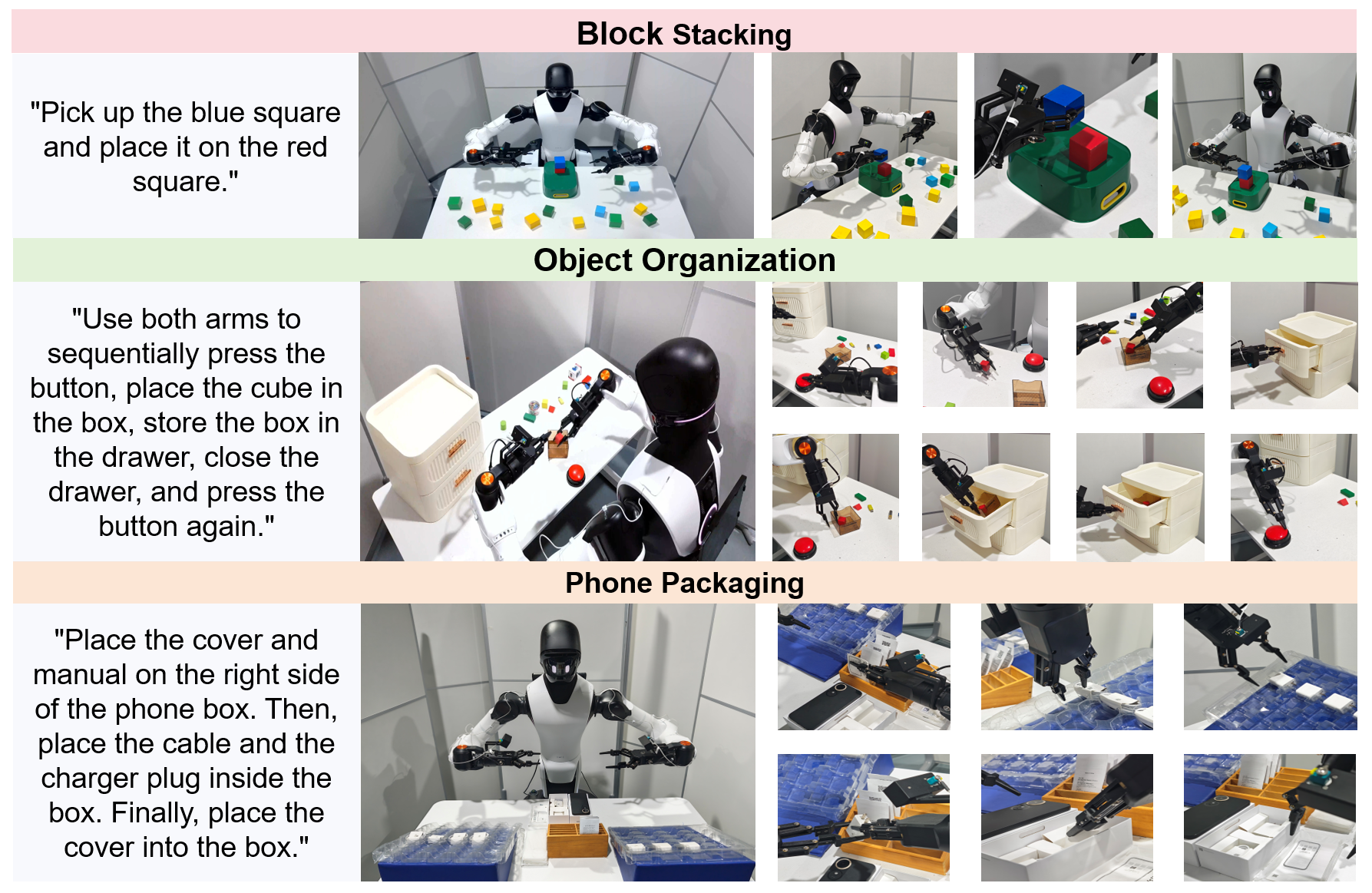}
  \caption{\textbf{Real-world manipulation tasks on the AGIBOT G2.}
  From top to bottom: block stacking, object organization, and phone packaging.
  Each task panel presents the language instruction (left), an overview of the workspace (center), and key frames illustrating the manipulation sequence (right).
  The tasks evaluate spatially precise placement, coordinated bimanual manipulation, and multi-step task execution.}
  \label{fig:real-world-setup}
  \vspace{-12pt}
\end{figure*}

\begin{table*}[!t]
  \vspace*{4pt}
  \centering
  \caption{\textbf{Success rate (\%) on the ten LIBERO-Mem tasks.}
  Best per column in \textbf{bold} (ties included).}
  \label{tab:libero-mem}
  \setlength{\tabcolsep}{2.8pt}
  {\scriptsize
  \begin{tabular}{@{}l*{11}{c}@{}}
    \toprule
    & \multicolumn{2}{c}{Single}
    & \multicolumn{4}{c}{Counting}
    & \multicolumn{4}{c}{Sequencing / Spatial}
    & \multirow{2}{*}{Avg.} \\
    \cmidrule(lr){2-3}\cmidrule(lr){4-7}\cmidrule(lr){8-11}
    Method
    & B$\rightarrow$P & Bt$\rightarrow$P
    & B$\times3$ & Bt$\times3$ & B$\times5$ & B$\times7$
    & Swap2 & Rot3 & Cs$\rightarrow$Bskt & Cs$+$E & \\
    \midrule
    $\pi_0$~\cite{pi0}                                     & 50 & 0 & 0 & \textbf{0} & 0 & 0 & 0 & 0 & 0 & 0 & 5.0 \\
    SlotVLA ($h=8$)~\cite{liberomem}                       & 50 & 0 & 0 & \textbf{0} & 0 & 0 & 0 & 0 & 0 & 0 & 5.0 \\
    MemoryVLA~\cite{memoryvla}                             & 55 & 0 & 35 & \textbf{0} & 5 & 0 & 0 & 0 & 0 & 0 & 9.5 \\
    $\pi_{0.5}^{\dagger}$~\cite{pi05}                      & \textbf{95} & \textbf{5} & 80 & \textbf{0} & 40 & 10 & \textbf{10} & \textbf{5} & \textbf{55} & \textbf{80} & 38.0 \\
    \midrule
    \textbf{\method{}}$^{\dagger}$ \textbf{(ours)}         & \textbf{95} & 0 & \textbf{90} & \textbf{0} & \textbf{70} & \textbf{40} & 5 & \textbf{5} & 25 & \textbf{80} & \textbf{41.0} \\

    \bottomrule
  \end{tabular}}
  \par\vspace{2pt}
  {\scriptsize B: bowl; Bt: bottle; P: plate; Cs: cream cheese; Bskt: basket; E: empty basket.
  $\times n$ denotes $n$ repetitions; Swap2 and Rot3 denote swapping two bowls and rotating three bowls.
  $^{\dagger}$ denotes models trained and evaluated by us;
  results without $^{\dagger}$ are reported baselines. Our models use $20$ trials per task.}
  \vspace{-12pt}
\end{table*}

\subsubsection{Evaluation on LIBERO}
We evaluate on all five LIBERO suites: libero-spatial, libero-object, libero-goal,
libero-10, and libero-90. We train a separate policy for each suite, using $40$k optimization
steps on the first four suites and $80$k steps on libero-90.
Both models are evaluated for $50$ trials per task with evaluation seed $7$, a replanning
interval of $8$ environment steps, and an environment rendering resolution of
$128\times128$.

\begin{table}[t]
  \centering
  \caption{\textbf{Success rate (\%) on the five MIKASA-Robo tasks.}
  Best per column in \textbf{bold} (ties included).}
  \label{tab:mikasa-robo}
  \setlength{\tabcolsep}{2.0pt}
  {\scriptsize
  \begin{tabular}{@{}lcccccc@{}}
    \toprule
    Method
    & \shortstack{ShellGame\\Touch}
    & \shortstack{Intercept\\Medium}
    & \shortstack{Remember\\Color3}
    & \shortstack{Remember\\Color5}
    & \shortstack{Remember\\Color9}
    & Avg. \\
    \midrule
    SpatialVLA~\cite{spatialvla}                           & 23 & 27 & 27 & 17 & 11 & 21.0 \\
    OpenVLA-OFT~\cite{openvlaoft}                          & 47 & 14 & \textbf{59} & 16 & 6 & 28.4 \\
    $\pi_0$~\cite{pi0}                                     & 33 & \textbf{42} & 35 & 22 & 15 & 29.4 \\
    $\pi_{0.5}^{\dagger}$~\cite{pi05}                      & 78 & 22 & 28 & 22 & 12 & 32.4 \\
    MemoryVLA~\cite{memoryvla}                             & \textbf{88} & 24 & 44 & 30 & 20 & 41.2 \\
    \midrule
    \textbf{\method{}}$^{\dagger}$ \textbf{(ours)}         & 86 & 38 & 46 & \textbf{32} & \textbf{24} & \textbf{45.2} \\
    \bottomrule
  \end{tabular}}
  \par\vspace{2pt}
  {\scriptsize $^{\dagger}$ denotes models trained and evaluated by us;
  results without $^{\dagger}$ are taken from MemoryVLA~\cite{memoryvla}. Our models use $100$ trials per task.}
\end{table}

\begin{table}[t]
  \centering
  \caption{\textbf{Success rate (\%) on SimplerEnv Bridge.}
  Best per column in \textbf{bold} (ties included).}
  \label{tab:simplerenv-bridge}
  \setlength{\tabcolsep}{2.2pt}
  {\footnotesize
  \begin{tabular*}{\columnwidth}{@{\extracolsep{\fill}}lccccc@{}}
    \toprule
    Method
    & \shortstack{Spoon on\\towel}
    & \shortstack{Carrot on\\plate}
    & \shortstack{Stack\\cube}
    & \shortstack{Eggplant in\\basket}
    & Avg. \\
    \midrule
    OpenVLA~\cite{openvla}                                 & 4.2 & 0.0 & 0.0 & 12.5 & 4.2 \\
    SpatialVLA~\cite{spatialvla}                           & 16.7 & 25.0 & 29.2 & \textbf{100.0} & 42.7 \\
    $\pi_{0.5}^{\dagger}$~\cite{pi05}                      & \textbf{83.3} & 37.5 & 45.8 & \textbf{100.0} & 66.7 \\
    MemoryVLA~\cite{memoryvla}                             & 75.0 & \textbf{75.0} & 37.5 & \textbf{100.0} & 71.9 \\
    \midrule
    \textbf{\method{}}$^{\dagger}$ \textbf{(ours)}         & 79.2 & 62.5 & \textbf{54.2} & \textbf{100.0} & \textbf{74.0} \\
    \bottomrule
  \end{tabular*}}
  \par\vspace{2pt}
  {\scriptsize $^{\dagger}$ denotes models trained and evaluated by us;
  results without $^{\dagger}$ are taken from the cited works. Our models use $24$ trials per task.}
   \vspace{-12pt}
\end{table}

As shown in Table~\ref{tab:main}, \method{} achieves an average success rate of
$97.8\%$ across the five LIBERO suites, compared with $96.6\%$ for \base{} and
$96.5\%$ for MemoryVLA. We observe higher success rates than \base{} on four of
the five suites, with the largest differences on libero-goal ($97.6\%$ vs.\ $94.6\%$)
and libero-10 ($97.6\%$ vs.\ $95.4\%$).


\subsubsection{Evaluation on LIBERO-Mem}
\label{sec:exp:libero-mem}
LIBERO-Mem~\cite{liberomem} evaluates history-dependent manipulation across ten tasks.
We train jointly across these tasks for $120$k optimization steps using
$961$ demonstrations retained after replay refinement. We evaluate both models for
$20$ trials per task and replan every $8$ environment steps.

As shown in Table~\ref{tab:libero-mem}, \method{} achieves an average success rate of
$41.0\%$, compared with $38.0\%$ for \base{}. The largest gains occur on the repeated-placement
tasks B$\times5$ and B$\times7$, each improving by $30$ percentage points over \base{}.

\subsubsection{Evaluation on MIKASA-Robo}
\label{sec:exp:mikasa-robo}
We evaluate on five memory-intensive manipulation tasks from MIKASA-Robo~\cite{mikasa}.
We train jointly across these tasks for $40$k optimization steps using
$250$ demonstrations per task. Both models are evaluated for $100$ trials per task,
with a replanning interval of $8$ environment steps.

Table~\ref{tab:mikasa-robo} reports an average success rate of $45.2\%$ for \method{},
exceeding \base{} and MemoryVLA by $12.8$ and $4.0$ percentage points, respectively.
We observe higher success rates than \base{} on all five tasks.

\subsubsection{Evaluation on SimplerEnv Bridge}
\label{sec:exp:simplerenv-bridge}
We train on BridgeData V2~\cite{bridgedatav2} for $60$k optimization steps and evaluate
both models on four WidowX tasks in SimplerEnv~\cite{simplerenv}, with $24$ trials per task. Both models
omit proprioceptive state and replan at every environment step, applying weighted smoothing
over the seven most recent action predictions with a coefficient of $0.1$.

As shown in Table~\ref{tab:simplerenv-bridge}, \method{} achieves an average success rate of
$74.0\%$, compared with $71.9\%$ for MemoryVLA. On Stack Cube, \method{} reaches $54.2\%$
versus $37.5\%$ for MemoryVLA, an improvement of $16.7$ percentage points.

\subsection{Real-World Experiments}

\label{sec:exp:real}


We conduct real-world experiments on three tasks: block stacking, object organization, and
phone packaging (Fig.~\ref{fig:real-world-setup}). All experiments use an AGIBOT G2 dual-arm
robot with one fixed head-view RGB camera and two wrist-mounted RGB cameras, one on each arm.
The recorded RGB images have resolutions of $640\times400$ pixels for the head camera
and $640\times528$ pixels for each wrist camera (width $\times$ height), before preprocessing.

\begin{table}[t]
  \centering
  \caption{\textbf{Success rate (\%) on real-world manipulation tasks.}
  Best result in each completed column is in \textbf{bold}.}
  \label{tab:real-world}
  \setlength{\tabcolsep}{3pt}
  {\footnotesize
  \begin{tabular*}{\columnwidth}{@{\extracolsep{\fill}}lcccc@{}}
    \toprule
    & \multicolumn{4}{c}{SR (\%) $\uparrow$} \\
    \cmidrule(lr){2-5}
    Method
    & \shortstack{Block\\Stacking}
    & \shortstack{Object\\Organization}
    & \shortstack{Phone\\Packaging}
    & Avg. \\
    \midrule
    $\pi_{0.5}$~\cite{pi05}
      & 41.7 & 55.0 & 10.0 & 35.6 \\
    Spatial Forcing~\cite{spatialforcing}
      & 58.3 & 40.0 & 10.0 & 36.1 \\
    MemoryVLA~\cite{memoryvla}
      & 88.3 & 35.0 & 15.0 & 46.1 \\
    \midrule
    \textbf{\method{}} \textbf{(ours)}
      & \textbf{100.0} & \textbf{85.0} & \textbf{45.0} & \textbf{76.7} \\
    \bottomrule
  \end{tabular*}}
   \vspace{-12pt}
\end{table}

\subsubsection{Block Stacking}
\label{sec:exp:real-stacking}
This task evaluates precise placement at varying target heights: the robot must place a blue
block stably on top of a red block. The demonstrations cover five configurations, in which
different support objects position the red block at different heights; the training
dataset contains $817$ demonstrations, on which we train \method{} for $30$k optimization steps. We evaluated 3 target-height configurations with $20$
trials per configuration for each method. A trial is
successful only if the blue block remains stably on the red block after release.

Table~\ref{tab:real-world} shows that \method{} achieves success $100.0\%$ in all $60$
trials, outperforming \base{} by $58.3$ percentage points, Spatial Forcing by $41.7$
percentage points and MemoryVLA by $11.7$ percentage points. Inspection of the evaluation
records indicates that \base{} and Spatial Forcing commonly fail through execution hesitation
or object drops during grasping and placement. MemoryVLA substantially reduces hesitation
failures, whereas \method{} completes every evaluated trial successfully.

\subsubsection{Object Organization}
\label{sec:exp:real-organization}
Success requires completing the organization sequence in Fig.~\ref{fig:real-world-setup},
beginning with a right-arm button press and ending with a left-arm button press.
We train \method{} on $237$ demonstrations for $20$k
optimization steps and evaluate each method for $20$ trials.

As shown in Table~\ref{tab:real-world}, \method{} achieves an $85\%$ success rate on object organization, outperforming \base{}, Spatial Forcing, and MemoryVLA by $30$, $45$, and $50$ percentage points, respectively. For \base{}, failures mainly involve unsuccessful block grasps and repeated left-arm button presses at the final step. Spatial Forcing exhibits the same failure modes, as well as repeated right-arm button presses at the initial step. MemoryVLA primarily fails to grasp the block, with repeated button pressing also observed. These observations highlight challenges in both precise grasping and tracking completed task steps.

\subsubsection{Phone Packaging}
\label{sec:exp:real-packaging}
The task requires sequentially packing a phone case and accessories, including a bimanual
charger handover and final placement of the phone case into the box (Fig.~\ref{fig:real-world-setup}). We train
\method{} on $203$ demonstrations for $40$k optimization steps and evaluate each method for $20$ trials.

\method{} achieves $45.0\%$ success, compared with $10.0\%$ for \base{} and Spatial Forcing and $15.0\%$ for MemoryVLA (Table~\ref{tab:real-world}). \base{} and Spatial Forcing mainly fail to grasp the phone case or charger; \base{} also exhibits prolonged hesitation during charger handover. Neither MemoryVLA nor \method{} incurs hesitation-related failures in these trials. Their remaining failures involve charger grasping or placement for MemoryVLA, and charger grasping or final phone-case placement for \method{}. These observations suggest that memory may ease subtask transitions, while precise grasping and placement remain challenging.

\subsection{Ablation Studies}
\label{sec:exp:ablation}
We ablate spatial tokenization, temporal memory, retrieval-bias strength, and the number of \geo{} input frames on real-world block stacking. Success rates are averaged over three target-height configurations, with $20$ trials per configuration. All other settings are held fixed within each study.

\subsubsection{Spatial Information and Memory}
\label{sec:exp:ablation-components}
We compare full \method{} with two variants: w/o memory removes the token memory bank and
retrieval while retaining spatial scene information, and w/o spatial removes the spatial-scene
information branch while retaining memory over 2D visual tokens. As shown in Table~\ref{tab:ablation}(a),
removing memory decreases the success rate from $100.0\%$ to $95.0\%$, while removing the
spatial-scene-information branch decreases it to $96.7\%$. The evaluation records further show distinct failure
patterns: the w/o-memory variant exhibits more execution hesitation, whereas the w/o-spatial
variant produces empty grasps and placement errors. These complementary failure modes indicate
that both temporal memory and spatial scene information contribute to reliable placement.


\begin{table}[t]
\vspace*{4pt}
  \centering
  \caption{\textbf{Ablation studies and design analysis on real-world block stacking.}
  Best result in each panel is in \textbf{bold}.}
  \label{tab:ablation}
  \setlength{\tabcolsep}{3pt}
  {\footnotesize
  \begin{tabular*}{\columnwidth}{@{\extracolsep{\fill}}lcccc@{}}
    \toprule
    \multicolumn{5}{@{}l}{\textbf{(a) Component ablation}} \\
    \midrule
    Method
      & $\pi_{0.5}$
      & \shortstack{\textbf{\method{}} \\ \textbf{(ours)}}
      & \shortstack{w/o\\memory}
      & \shortstack{w/o\\spatial} \\
    \midrule
    SR (\%) $\uparrow$
      & 41.7
      & \textbf{100.0}
      & 95.0
      & 96.7 \\
    \bottomrule
  \end{tabular*}
  \par\vspace{4pt}

  \begin{tabular*}{\columnwidth}{@{\extracolsep{\fill}}lcccccc@{}}
    \toprule
    \multicolumn{7}{@{}l}{\textbf{(b) Retrieval-bias scale}} \\
    \midrule
    Bias scale $\beta$ & $0$ & $0.05$ & $0.3$ & $0.5$ & $0.8$ & $1.0$ \\
    \midrule
    SR (\%) $\uparrow$ & 80.0 & 71.7 & 71.7 & \textbf{100.0} & 71.7 & 81.7 \\
    \bottomrule
  \end{tabular*}

  \par\vspace{4pt}

  \begin{tabular*}{\columnwidth}{@{\extracolsep{\fill}}lccccc@{}}
    \toprule
    \multicolumn{6}{@{}l}{\textbf{(c) Number of input frames for \geo{}}} \\
    \midrule
    Input frames $K$ & $1$ & $2$ & $3$ & $4$ & $5$ \\
    \midrule
    SR (\%) $\uparrow$ & 88.3 & 93.3 & 85.0 & \textbf{100.0} & 86.7 \\
    \bottomrule
  \end{tabular*}

  }
  \par\vspace{2pt}
  {\scriptsize Success rates are averaged over three target-height configurations,
  with $20$ trials per configuration. Default settings: $\beta=0.5$ and $K=4$.
  All other settings are held fixed within each study.\par}
  \vspace{-12pt}
\end{table}

\subsubsection{Retrieval Bias Strength}
\label{sec:exp:ablation-bias}
We vary the retrieval-bias scale in Eq.~\eqref{eq:retrieval-bias} while fixing the temperature at $\tau_{\mathrm r}=1$. Setting $\beta=0$ disables the score-induced bias without removing memory retrieval.

Table~\ref{tab:ablation}(b) shows that success varies non-monotonically with bias strength, with the highest observed success rate of $100.0\%$ at $\beta=0.5$, compared with $80.0\%$ without bias. At $\beta=1.0$, success is $81.7\%$, while the remaining tested nonzero scales fall below the unbiased setting. The benefit therefore depends on the bias scale rather than increasing with it. We use $\beta=0.5$ as the default.

\subsubsection{Input Frames for \geo{}}
\label{sec:exp:ablation-temporal}
We vary the number of head-camera frames supplied to \geo{} while keeping the sampling scheme and memory capacity fixed. The current frame is always included; $K=1$ removes supporting frames but retains spatial tokens and memory.

Table~\ref{tab:ablation}(c) shows a non-monotonic response, with the highest success rate of $100\%$ at $K=4$. Additional frames provide more observations for current-scene spatial feature extraction but increase inference latency, so more input frames do not necessarily improve closed-loop performance.
\section{CONCLUSION}
\label{sec:conclusion}


We presented \method{}, a VLA framework that incorporates pretrained visual geometry features
into temporal memory, allowing past scene structure and appearance to inform current actions.
Its joint visual-spatial retrieval mechanism uses a mathematically grounded relevance--non-redundancy score, computed without additional learned parameters, to guide cross-attention toward relevant and complementary history. \method{} improves average success over \base{}
on four simulation benchmarks and increases success by $30.0$--$58.3$ percentage points across
three real-world tasks. Component ablations on block stacking further suggest that both spatial
information and temporal memory contribute to reliable placement.

Our approach nevertheless relies on a pretrained 3D foundation model whose features do not
always yield high-quality spatial tokens. In addition, the current memory bank maintains only a
short temporal window, which limits its benefits on substantially longer-horizon tasks. Future
work will improve spatial scene perception across a broader range of tasks and environments,
and investigate learnable memory mechanisms that can retain and retrieve information over
longer time horizons.


\bibliographystyle{IEEEtran}
\bibliography{refs}

\ifappendix
  \appendix
  \input{sections/07_appendix}
\fi

\addtolength{\textheight}{0cm}

\end{document}